\documentclass[letterpaper, 10pt, conference]{ieeeconf}

\IEEEoverridecommandlockouts
\usepackage[utf8]{inputenc}
\usepackage[T1]{fontenc}
\usepackage{amsmath}
\usepackage{amssymb}
\usepackage{bm}

\title{\LARGE \bf
GraspSense: Physically Grounded Grasp and Grip Planning for a Dexterous Robotic Hand via Language-Guided Perception and Force Maps
}

\author{%
    Elizaveta Semenyakina*,
    Ivan Snegirev*,
    \thanks{*These authors contributed equally to this work.}
    Mariya Lezina, \\
    Miguel Altamirano Cabrera,
    Safina Gulyamova
    and Dzmitry Tsetserukou%
    \thanks{The authors are with the Intelligent Space Robotics Laboratory,
    Skolkovo Institute of Science and Technology. 
    \{elizaveta.semenyakina, ivan.snegirev, mariya.lezina,
    m.altamirano, safina.gulyamova, d.tsetserukou\}@skoltech.ru}
}

\usepackage{graphicx}
\begin{document}

\maketitle
\thispagestyle{empty}
\pagestyle{empty}

\begin{abstract}
Dexterous robotic manipulation requires more than geometrically valid grasps: it demands physically grounded contact strategies that account for the spatially non-uniform mechanical properties of the object. However, existing grasp planners typically treat the surface as structurally homogeneous, even though contact in a weak region can damage the object despite a geometrically perfect grasp.
We present a pipeline for grasp selection and force regulation in a five-fingered robotic hand, based on a map of locally admissible contact loads. From an operator command, the system identifies the target object, reconstructs its 3D geometry using SAM3D, and imports the model into Isaac Sim. A physics-informed geometric analysis then computes a force map that encodes the maximum lateral contact force admissible at each surface location without deformation.
Grasp candidates are filtered by geometric validity and task-goal consistency. When multiple candidates are comparable under classical metrics, they are re-ranked using a force-map-aware criterion that favors grasps with contacts in mechanically admissible regions. An impedance controller scales the stiffness of each finger according to the locally admissible force at the contact point, enabling safe and reliable grasp execution.
Validation on paper, plastic, and glass cups shows that the proposed approach consistently selects structurally stronger contact regions and keeps grip forces within safe bounds. In this way, the work reframes dexterous manipulation from a purely geometric problem into a physically grounded joint planning problem of grasp selection and grip execution for future humanoid systems.

\end{abstract}

\section{Introduction}

\begin{figure}[t]
    \centering
    \includegraphics[width=\linewidth]{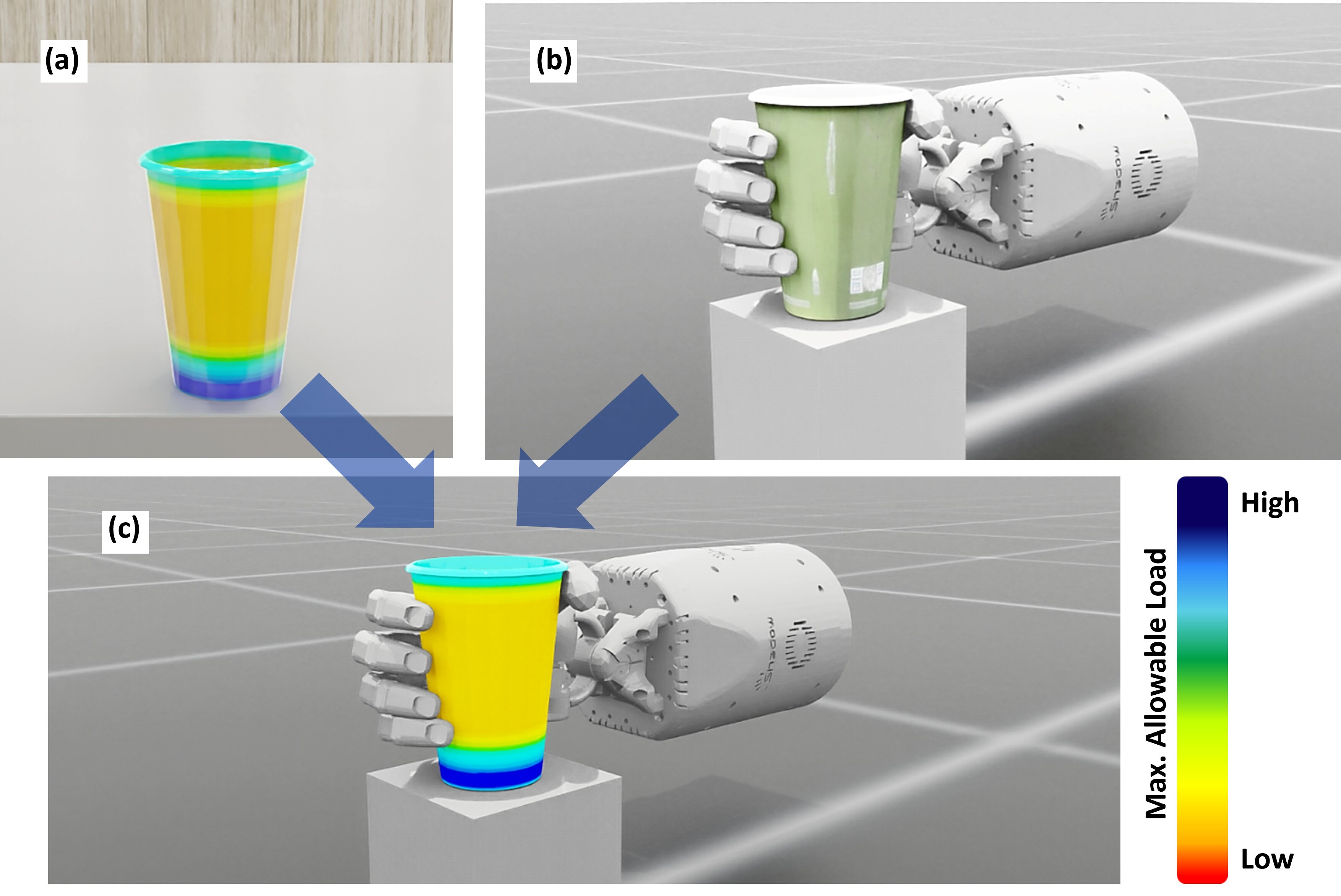}
    \caption{Illustration of the force-map-aware grasp planning approach
on a paper cup.
(a)~The reconstructed 3D object model with the computed force map: colour encodes the locally admissible lateral contact force at each
surface point (blue – high admissible load; red
– low admissible load).
(b)~Grasp executed using geometric feasibility criteria alone,
without force-map information.
(c)~Grasp executed with force-map-aware re-ranking: contacts are
directed toward the mechanically admissible mid-body and rim zones,
and per-finger stiffness is scaled to the locally admissible force,
ensuring reliable retention without structural overloading.}
\label{fig:teaser}
\vspace{-0.3cm}
\end{figure}

Robust physical interaction with everyday objects remains one of the central open problems in robotics.
While parallel-jaw grippers have enabled significant progress in industrial automation, their inherent simplicity limits applicability to objects of varying shape, fragility, and material. Multi-fingered dexterous hands, capable of replicating the richness of human grasp repertoires, have long been recognized as a key enabler for general-purpose manipulation~\cite{billard2019}. Yet the very flexibility that makes such hands powerful also makes grasp planning substantially harder: for a given object, a five-fingered hand can realize thousands of kinematically feasible configurations, many of which are equivalent under classical stability metrics.

The dominant paradigm in grasp planning, from analytic force-closure methods to large-scale learned approaches such as DexGraspNet~\cite{dexgraspnet}, evaluates candidate grasps primarily in terms of geometric and kinematic properties: stability, reachability, and absence of collisions. These criteria, while necessary, are not sufficient. In physical manipulation, the quality of a grasp is ultimately determined not only by the configuration of the hand but also by the forces the hand exerts on the object at contact. Two grasps that are geometrically equivalent may differ dramatically in
their mechanical admissibility: one may engage a structurally robust zone of the object surface, while the other may apply load to a region prone to deformation or failure.
For fragile, thin-walled or heterogeneous materials, it determines whether the item will withstand the grip.

This paper addresses the gap between geometric grasp planning and physically informed manipulation. We argue that, for a multi-fingered hand, grasp selection and grip force regulation cannot be treated as independent subproblems: they must be solved jointly, with explicit reference to the spatially non-uniform mechanical properties of the target object.
We introduce a pipeline that unifies language-level task understanding, object reconstruction and material inference, physics-informed structural load analysis, dexterous grasp generation and ranking, and compliant grip execution within a single coherent framework.

The key contributions of this work are as follows:
\begin{enumerate}
  \item A \emph{force map construction module} that estimates spatially distributed admissible contact loads for a reconstructed 3D object model via a physics-informed geometric approximation of local wall thickness, providing a per-surface-region bound on mechanically safe contact forces.

  \item A \emph{physically grounded grasp selection criterion} that extends classical multi-criteria ranking (stability, reachability, collision avoidance) with a force-map-aware re-ranking stage, selecting among geometrically equivalent candidates the one whose contact regions are most mechanically admissible.

  \item An \emph{impedance-based grip execution strategy} that scales per-finger stiffness to the locally admissible force at each contact point, implementing a spatially non-uniform grip strategy that is safe with respect to object structure while reliably retaining the object.

  \item Full pipeline integration, from natural language command to grip execution, in Isaac Sim, demonstrated on cup-like objects of three distinct materials: paper, plastic, and glass.
\end{enumerate}

\section{Background and Related Work}

\subsection{Dexterous Grasp Generation and Ranking}

Grasp synthesis for multi-fingered hands encompasses both analytic approaches, which formulate grasp quality in terms of contact stability conditions and quantitative grasp quality metrics, and data-driven methods that learn to generate grasp candidates from object geometry.
DexGraspNet~\cite{dexgraspnet} establishes a large-scale
simulation-based benchmark and generative model for dexterous grasps
using the Shadow Hand~\cite{shadowhand}; UniDexGrasp~\cite{unidexgrasp} and
AnyDexGrasp~\cite{anydexgrasp} extend this toward universal grasp
generation across hand morphologies.
When multiple candidates are geometrically valid, a ranking stage is
required. GRaCE~\cite{grace} introduces a probabilistic framework for
multi-criteria ranking that balances stability, kinematic feasibility,
and task-relevant objectives via hierarchical logic and a
rank-preserving utility function. Our work extends this by
incorporating local force admissibility – derived from structural
analysis – as an additional ranking criterion, enabling selection
among geometrically equivalent candidates on the basis of mechanical
safety.

\subsection{Force-Aware and Compliant Manipulation}

Contact forces have long been central to grasp quality, from classical
contact stability analysis to more recent adaptive manipulation methods
that incorporate force feedback during execution~\cite{tian2024}. Impedance control offers
compliant and physically consistent force regulation~\cite{hogan1985};
our grip execution module builds on this foundation, but grounds the
impedance setpoints in force-map-derived bounds specific to the
object's local structure rather than generic compliance parameters.
Prior structural analysis in grasping has focused on modelling contact
mechanics or force–deformation relationships in deformable-object
manipulation~\cite{dharbaneshwer2021}. We use structural reasoning
differently: to pre-compute a spatially distributed map of
mechanically admissible surface loads that directly informs both grasp
selection and force control.

\subsection{Language-Guided Manipulation}

Large language models (LLM) are increasingly used in robotic manipulation for high-level task planning and language-guided grasping~\cite{codeaspolicies,langrasp,graspasyousay,tang2023}. In our pipeline, Qwen~\cite{qwen} parses operator commands to extract the target object, action type, and interaction mode, while the resulting semantic context – particularly material type – is propagated to the structural analysis stage to condition the material model.

\subsection{Perception and 3D Reconstruction for Manipulation}

SAM~\cite{sam} enables zero-shot object segmentation from 2D images, while YOLO-World~\cite{yoloworld} provides open-vocabulary object localization suitable for downstream reconstruction. Reconstructed object models can then be imported into Isaac Sim for high-fidelity simulation of contact and force dynamics. Our pipeline combines YOLO-World detection, SAM segmentation, SAM3D reconstruction, and USD conversion into a unified perception front-end for structural analysis and grasp planning.

\subsection{Gap Addressed by This Work}

Despite the substantial progress reviewed above, no existing system
simultaneously addresses grasp generation, multi-criteria ranking, and
grip force regulation in a manner grounded in the spatially
heterogeneous mechanical properties of the target object.
Existing grasp planners treat object surfaces as mechanically uniform,
and force-control methods typically operate on generic compliance
setpoints unrelated to object structure.
Our work closes this gap by introducing the force map as a first-class
element of both grasp selection and grip execution, enabling a
transition from purely geometric manipulation planning to physically
grounded interaction planning.

\begin{figure*}[t]
    \centering
    \includegraphics[width=\textwidth]{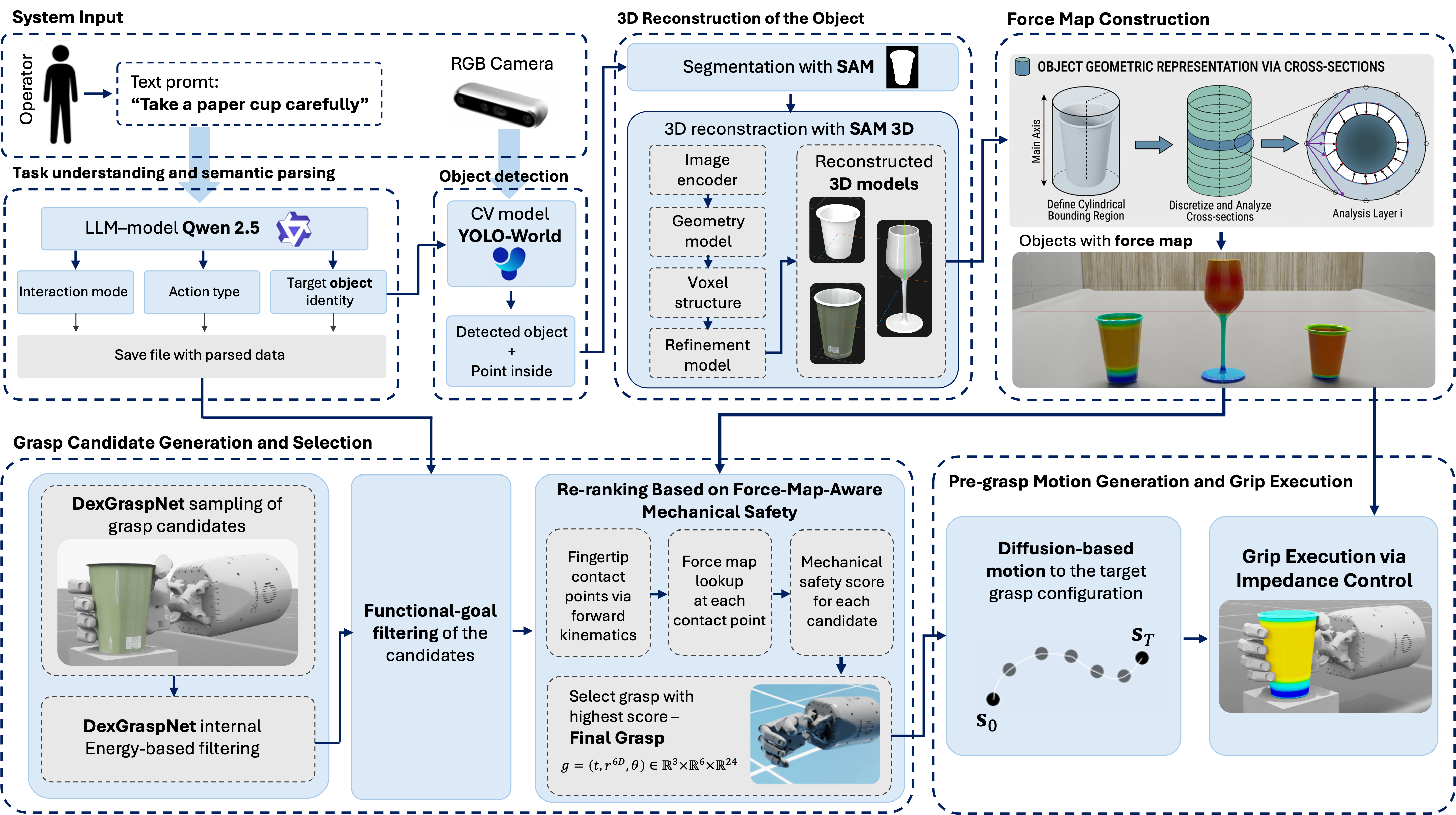}
    \caption{Overview of the proposed pipeline, comprising five stages:
task understanding and semantic parsing,
object perception and 3D reconstruction,
force map construction through physics-informed geometric approximation,
grasp candidate generation with functional-goal filtering and
force-map-aware re-ranking, and
pre-grasp motion generation and grip execution.}
\label{fig:pipeline}
\vspace{-0.3cm}
\end{figure*}

\section{System Overview and Pipeline}

An overview of the proposed system pipeline is shown in Fig.~\ref{fig:pipeline}. The proposed system transforms a natural language operator command into
a physically grounded grasp-and-grip plan for a five-fingered robotic
hand.
The pipeline is organized into five major stages:
(A)~task understanding and semantic parsing,
(B)~object perception and 3D reconstruction,
(C)~force map construction through physics-informed geometric approximation, 
(D)~grasp candidate generation, functional-goal filtering, and force-map-aware re-ranking, and (E)~pre-grasp motion generation and grip execution.

\subsection{Task Understanding via Language Model (Qwen)}

The pipeline is initiated by a natural language command issued by a
human operator as typed text or transcribed speech, parsed by
Qwen~\cite{qwen} in inference mode without fine-tuning.

The LLM extracts three components: (1) target object identity
($\mathcal{O} \in \{\text{paper cup, plastic cup, glass goblet}\}$),
passed to both the perception module and the structural analysis stage
to condition the material model; (2) action type
(pick up, pick and place, or hand over),
preserved as task-level context; and (3) interaction mode
(\textit{"gently"} / default / \textit{"firmly"}), mapped to a
normalized force scaling factor $\lambda \in [0,1]$.

The LLM output is the tuple $\mathcal{C} = (\mathcal{O}, a, \lambda)$,
which propagates downstream through the entire pipeline.

\subsection{Object Perception and 3D Reconstruction}

\subsubsection{Object Detection and Segmentation}

YOLO-World~\cite{yoloworld} performs open-vocabulary detection from the text query $\mathcal{O}$, returning a point inside the detected object. This point serves as a spatial prompt to SAM~\cite{sam}, which generates a pixel-accurate binary mask isolating the target from background clutter.
The stage is designed to operate robustly under cluttered industrial conditions, including partial occlusion and uncontrolled lighting.

\subsubsection{3D Reconstruction (SAM3D)}

After the mask is obtained, the original RGB image and the binary mask
are passed to SAM3D~\cite{sam3d} for object-centered 3D reconstruction.
The model uses both the visual appearance of the object and its precise
spatial boundaries in the image to infer a three-dimensional
representation. First, a visual encoder extracts features related to
the object's shape, texture, and structure. These features are then
used to form an intermediate sparse 3D structure, which serves as a
coarse approximation of the object's volume, silhouette, and overall
geometry. This representation is subsequently refined into a more
complete 3D model, from which different output formats can be decoded.
Depending on the decoder configuration, SAM3D can produce either a
Gaussian-based representation, a polygonal mesh, or both. In our
pipeline, the reconstructed object is exported as a mesh and converted
to \texttt{.glb} format for further use in Isaac Sim.

\subsubsection{GLB-to-USD Conversion and Scene Import}

The \texttt{.glb} mesh is converted to \texttt{.usd} via a custom
wrapper around Isaac Sim's conversion utility.
Three property groups are assigned: geometric normalization (mesh
rescaled to real-world reference dimensions); SDF (Signed Distance Field) mesh collider
(required for accurate contact-region localization in the force-map
stage); and material physics ($\rho$, $\mu_s$, $\mu_d$, $e_r$ from a
lookup table keyed to $\mathcal{O}$, with mass computed automatically).
The USD asset is inserted into the scene as a reference, with placement
validated automatically.

\subsection{Force Map Construction: Physics-Informed Geometric
Approximation}

The force map provides, for each point on the object surface, an
estimate of the \emph{locally admissible lateral contact force} – the maximum force that can be applied without causing irreversible
deformation of the object wall. The approach uses a discretized
geometric representation conceptually related to FEM (Finite Element Method) preprocessing,
but instead of solving a full PDE system it employs a lighter
\emph{physics-informed geometric approximation} grounded in local
object geometry and material properties.

\subsubsection{Principal Axis Decomposition}

PCA (Principal Component Analysis) is applied to the mesh vertices $\mathbf{V} = \{\mathbf{v}_i\}$
to obtain a principal coordinate frame
$(\mathbf{c}, \mathbf{e}_u, \mathbf{e}_v, \mathbf{e}_h)$,
where $\mathbf{e}_h$ is the principal elongation axis (coinciding with
the vertical symmetry axis for cup-like objects) and $\mathbf{e}_u$,
$\mathbf{e}_v$ span the cross-sectional plane.
All subsequent operations are performed in this local frame.

\subsubsection{Cylindrical Discretization and Wall-Thickness Estimation}

The lateral surface is discretized into $L$ height layers and
$A$ angular bins. Each cell $(l, a)$, where $l \in \{0,\ldots,L{-}1\}$
denotes the layer index and $a \in \{0,\ldots,A{-}1\}$ the angular bin,
is associated with a local wall thickness $\tau(l,a)$ estimated via
a two-stage raycasting procedure: an inward ray from outside the object
yields the outer-surface contact point $\mathbf{p}_1$ and its normal
$n_1$; a second ray along the in-plane projection of
$n_1$ yields the inner-wall point $\mathbf{p}_2$, giving
\[
\tau(l,a)=\|p_2-p_1\|.
\]
Multiple probing directions per angular cell improve robustness.
The raw thickness map is completed and smoothed along both axes to
yield a stable field~$\mathcal{T}$.

\subsubsection{Edge Refinement}

Near object boundaries, thickness measurements may drop artificially
because rays miss the inner wall as the object terminates - yet
real cups often have reinforced rims precisely there.
An adaptive refinement stage is triggered for these layers: angular
resolution is doubled, the chord set expanded, and neighboring
sublayers additionally sampled, allowing the algorithm to distinguish
genuinely thin regions from structurally reinforced edges.

\subsubsection{Admissible Force Computation}

A base admissible force $F_{\text{base}}(l,a)$ is computed from
$\mathcal{T}$, increasing with local wall thickness.
Locally reinforced layers $\mathcal{S}$ (local force maxima) spread a
Gaussian vertical bonus $b_s(l)$ to neighboring layers, reflecting the
stiffening influence of thicker wall sections.
After material scaling, the final force map is:
\[
F(l,a)=\min\!\left(
  \bigl(F_{\text{base}}(l,a)+{\textstyle\sum_{s\in\mathcal{S}}}b_s(l)\bigr)
  \cdot k_m,\; F_{\text{clamp}}\right),
\]
where $k_m$ is a material-dependent coefficient and $F_{\text{clamp}}$
is a hard safety ceiling.

\subsubsection{Projection onto Mesh Vertices}

$F \in \mathbb{R}^{L \times A}$ is projected onto mesh vertices via
bilinear interpolation in cylindrical coordinates. Per-vertex admissible
force values are stored directly in the USD model as a
vertex-interpolated primvar, enabling runtime lookup at any contact
point without additional geometric processing. A companion colour
primvar encodes the map as a heat-map gradient for visual inspection
in Isaac Sim.

\subsection{Grasp Candidate Generation, Functional-Goal Filtering,
and Force-Map-Aware Re-Ranking}

\subsubsection{Grasp Representation}

Each grasp candidate for the Shadow Hand is represented as:
\begin{equation}
g = (t,\; r^{6D},\; \theta)
\in \mathbb{R}^{3} \times \mathbb{R}^{6} \times \mathbb{R}^{24},
\end{equation}
where $t$ is the hand base position in the object frame,
$r^{6D}$ is the continuous 6D rotation
representation, and $\theta$ are the 24
target joint angles of the Shadow Hand.

\subsubsection{Candidate Generation via DexGraspNet}

Candidates are generated using DexGraspNet~\cite{dexgraspnet} via
gradient-based optimization of a composite energy function:
\begin{equation}
E(g) = E_\text{fc} + E_\text{pen} + E_\text{self} + E_\text{joint},
\end{equation}
where $E_\text{fc}$ is a differentiable force-closure term ensuring
stable contact closure; $E_\text{pen}$ penalizes hand-object penetration;
$E_\text{self}$ penalizes self-collisions; and $E_\text{joint}$
enforces joint-angle limits. Candidates are subsequently validated in
Isaac Gym; those failing the stability test are discarded, yielding a
feasible set $\mathcal{G} = \{g_1, \ldots, g_K\}$ satisfying all hard
physical constraints.

\subsubsection{Functional-Goal Filtering}

A functional criterion $c_F$, conditioned on scene observation
$\mathbf{o}$ and target task $y$, is evaluated for each
$\mathbf{g} \in \mathcal{G}$. Candidates not satisfying it are
discarded:
\begin{equation}
\mathcal{G}^* =
\{ g \in \mathcal{G} \mid P(c_F(g) \mid o, y)
\geq \tau_F \},
\end{equation}
where $\tau_F$ is the acceptance threshold. We use
TaskGrasp~\cite{taskgrasp} as the task-oriented evaluator for $c_F$.

\subsubsection{Force-Map-Aware Re-Ranking}

After functional filtering, $\mathcal{G}^*$ contains candidates that
are both physically feasible and functionally suitable, yet may remain
geometrically indistinguishable under classical metrics.
The force-map-aware re-ranking stage selects among them the candidate
whose contact regions are most mechanically admissible - the central
contribution of this work.

\textit{Contact region identification.}
For each $\mathbf{g}_i \in \mathcal{G}^*$, fingertip positions in the
object frame are obtained via forward kinematics applied to
$(\mathbf{t}_i, \mathbf{r}_i^{6D}, \boldsymbol{\theta}_i)$.
Each contact point $\mathbf{p}_{i,k}$ is mapped to the containing
triangle $\Delta_{i,k}$ of the object mesh.

\textit{Admissible force at contact.}
The admissible force at contact point $\mathbf{p}_{i,k}$ is computed
by barycentric interpolation over the triangle vertices:
\begin{equation}
F_{i,k} =
\lambda_{i,k}^{(1)} F\!\left(v_{i,k}^{(1)}\right) +
\lambda_{i,k}^{(2)} F\!\left(v_{i,k}^{(2)}\right) +
\lambda_{i,k}^{(3)} F\!\left(v_{i,k}^{(3)}\right),
\end{equation}
where $\lambda_{i,k}^{(j)}$ are the barycentric coordinates of
$\mathbf{p}_{i,k}$ within $\Delta_{i,k}$.

\textit{Force-map score.}
The mechanical quality of a candidate is evaluated as:
\begin{equation}
S_F(g_i) = \min_k F_{i,k} - \alpha \cdot \mathrm{Var}_k(F_{i,k}).
\label{eq:sf}
\end{equation}

The first term rewards candidates whose weakest contact point admits the
largest safe force.
The second term penalizes high variance across fingers, since uneven
load distribution increases localized overloading risk.
$\alpha \geq 0$ controls the safety vs.\ uniformity trade-off.

\textit{Interaction mode modulation.}
If the operator-specified $\lambda$ is small (delicate interaction),
candidates with any contact point satisfying $F_{i,k} < \lambda \cdot
F_{\max}$ are additionally discarded before re-ranking.

\textit{Final grasp selection.}
\begin{equation}
g^* = \arg\max_{g_i \in \mathcal{G}^*} S_F(g_i).
\end{equation}
The output $g^* = (t^*, r^{6D*}, \theta^*)$ fully specifies hand position, orientation, and
finger configuration for downstream execution.


\subsection{Pre-Grasp Motion Generation and Grip Execution}

\subsubsection{Pre-Grasp Motion Generation}

A pre-grasp configuration is constructed by retracting the hand
along the palm approach axis with fingers in the open pose
$\boldsymbol{\theta}_\text{open}$, defining the transition:
\begin{equation}
s_0 =
(t_\text{pre},\, r^{6D*},\, \theta_\text{open})
\;\longrightarrow\;
s_T =
(t^*,\, r^{6D*},\, \theta^*).
\end{equation}
The approach trajectory is generated by a diffusion-based motion
planner conditioned on both endpoints and
the object geometry, and executed via a joint-space PD (Proportional-Derivative) controller
at 60~Hz.

\subsubsection{Grip Execution via Impedance Control}

Once contact is established at $\mathbf{s}_T$, the pipeline
transitions to grip execution, targeting sufficient retention force
without exceeding the locally admissible bounds from the force map.

\textit{Contact state estimation:}
At each control step, the physics engine provides per-finger contact
points $\mathcal{C}_k = \{c_{k,j}\}$ with normals $n^w_{k,j}$ and
impulses $J^w_{k,j}$. The admissible force $F^{\max}_{k,j}$ is looked
up via nearest-vertex query; the actual normal force
$F^{\text{est}}_{k,j}$ is estimated from the impulse normal component.
For each finger, the most critical contact point $j^*(k)$ is selected
as the one with the highest load ratio $F^{\text{est}}_{k,j} /
F^{\max}_{k,j}$, yielding the per-finger control parameters:
\begin{equation}
F^{\max}_k = F^{\max}_{k,j^*(k)}, \quad
n_k = \frac{n^w_{k,j^*(k)}}{\|n^w_{k,j^*(k)}\|}, \quad
f_k(t) = F^{\text{est}}_{k,j^*(k)}\,n_k.
\end{equation}

\textit{Adaptive impedance controller.}
Grip is implemented via Cartesian impedance control~\cite{hogan1985}:
\begin{equation}
M_k \Delta\ddot{x}_k + D_k \Delta\dot{x}_k + K_k \Delta x_k
= f_k(t) - f^{\text{des}}_k(t).
\end{equation}
Finger Cartesian targets are mapped to joint commands via
least-squares inverse kinematics. Normal stiffness is set adaptively from the
force map and interaction mode $\lambda$:
\begin{equation}
k_{n,k}(t) = \gamma_n \,\frac{F^{\text{tar}}_k(t)}{\delta_{n,\max}},
\end{equation}
with damping co-tuned to maintain a consistent transient response:
\begin{equation}
d_{n,k}(t) = 2\zeta_n \sqrt{m_{n,k}\, k_{n,k}(t)}.
\end{equation}
Fingers on weaker surface regions operate at lower stiffness;
fingers on stronger regions may exert greater force – implementing
the spatially non-uniform grip strategy that is the central
objective of the pipeline.

\section{Experimental Evaluation}

All experiments are conducted in Isaac Sim using a Shadow Hand model
with $K = 50$ DexGraspNet candidates per object.

\subsection{Experiment 1: Force-Map-Aware Grasp Selection}

Three objects are evaluated: a paper cup (7.6~g), a plastic cup
(1.9~g), and a glass goblet (155~g). Table~\ref{tab:force_map_zones}
reports the admissible lateral force per structural zone. For the
goblet, the stem is the structurally thicker lower section and the
bowl is the thin-walled upper part.

\begin{table}[h]
\caption{Admissible Lateral Force per Structural Zone.}
\label{tab:force_map_zones}
\centering\small\renewcommand{\arraystretch}{1.15}
\begin{tabular}{l c c c}
\hline
\textbf{Object} & \textbf{Zone 1 [N]} & \textbf{Zone 2 [N]} & \textbf{Base [N]} \\
\hline
Paper cup    & rim: 41.2 & wall: 4.8 & 88.8 \\
Plastic cup  & rim: 3.7  & wall: 0.3 &  8.0 \\
Glass goblet & bowl: 69  & stem: 565 & 989  \\
\hline
\end{tabular}
\end{table}

After feasibility and functional-goal filtering, the top-$M{=}5$
candidates are passed to two selection strategies:
Baseline – highest utility score $U(\mathbf{g})$, no
force-map information; Ours – candidate maximising
$S_F(\mathbf{g}_i)$ (Eq.~6).
We report contact zone, $F^*_{\min}$, safety margin
$F^*_{\min}/f_{\max}$, force-bound violations, and grasp success
(object held for 3~s).

\subsubsection{Results}

Table~\ref{tab:exp1} shows that the proposed method consistently
selects structurally reinforced zones – the rim for cups, the stem
for the goblet – improving $F^*_{\min}$ by $8.6\times$,
$12.3\times$, and $8.2\times$ respectively, with no force-bound
violations in any configuration.

\begin{table}[h]
\caption{Grasp Selection Results. Margin $= F^*_{\min}/f_{\max}$;
Viol.\ $=$ fingers with $f_k{>}F^*_k$.}
\label{tab:exp1}
\centering\small\setlength{\tabcolsep}{4pt}\renewcommand{\arraystretch}{1.15}
\begin{tabular}{llcccc}
\hline
\textbf{Object} & \textbf{Method} & \textbf{Zone}
  & $\boldsymbol{F^*_{\min}}$ & \textbf{Margin} & \textbf{Viol.} \\
\hline
Paper cup   & Baseline & wall &  4.8 & 1.4 & 1/5 \\
            & Ours     & rim  & 41.2 & 7.8 & 0/5 \\
\hline
Plastic cup & Baseline & wall &  0.3 & 1.2 & 2/5 \\
            & Ours     & rim  &  3.7 & 5.9 & 0/5 \\
\hline
Goblet      & Baseline & bowl & 69.0 & 2.1 & 1/5 \\
            & Ours     & stem & 565  &12.4 & 0/5 \\
\hline
\end{tabular}
\end{table}

\subsubsection{Qualitative Illustration: Glass Goblet}

\begin{figure}[h]
  \centering
  \includegraphics[width=0.49\columnwidth]{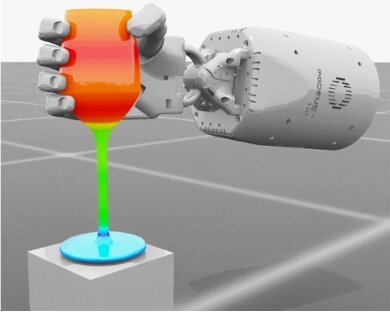}
  \hfill
  \includegraphics[width=0.49\columnwidth]{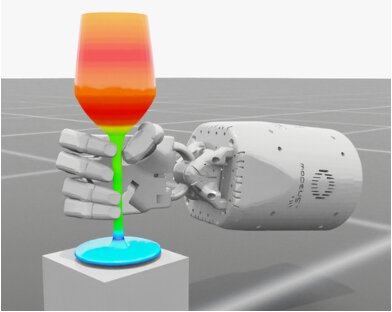}
  \caption{Two geometrically equivalent grasps on a glass goblet
  (blue $\rightarrow$ low; red $\rightarrow$ high admissible force).
  \textit{Left (rejected):} bowl contacts; $F^*_{\min}{=}69$~N.
  \textit{Right (selected):} stem contacts; $F^*_{\min}{=}565$~N.}
  \label{fig:goblet_grasps}
  \vspace{-0.5cm}
\end{figure}

The goblet illustrates the key insight most clearly: admissible force
differs by ${\sim}8\times$ between bowl (69~N) and stem (565~N), yet
the two grasps shown in Fig.~\ref{fig:goblet_grasps} differ by less
than 3\% on all classical metrics. Force-map-aware re-ranking
unambiguously selects Grasp~A (stem, $F^*_{\min}{=}565$~N) over
Grasp~B (bowl, $F^*_{\min}{=}69$~N).

\subsection{Experiment 2: Grip Force Regulation via Impedance Control}

We use the plastic cup (1.9~g), whose low admissible wall load (0.3~N)
makes the safe/damaging boundary most informative.
The force-map-selected grasp is fixed throughout; three controller
conditions are evaluated: \textit{under-stiffness}
($0.3{\cdot}K_{\text{base}}$, force map ignored) – insufficient
friction, object slips; over-stiffness
($2.0{\cdot}K_{\text{base}}$, force map ignored) – object retained
but $F^*_k$ exceeded with visible wall deformation; ours
($K_k = K_{\text{base}}{\cdot}F^*_k/F_{\max}{\cdot}\lambda$,
$\lambda{=}0.7$) – per-finger calibration, retained without
violation. Each condition is visualised as an Isaac Sim snapshot with
per-fingertip force vectors: length encodes magnitude, colour encodes
$f_k/F^*_k$ (green: safe; yellow: near-limit; red: violated).

\begin{figure}[h]
  \centering
  \includegraphics[width=0.32\columnwidth]{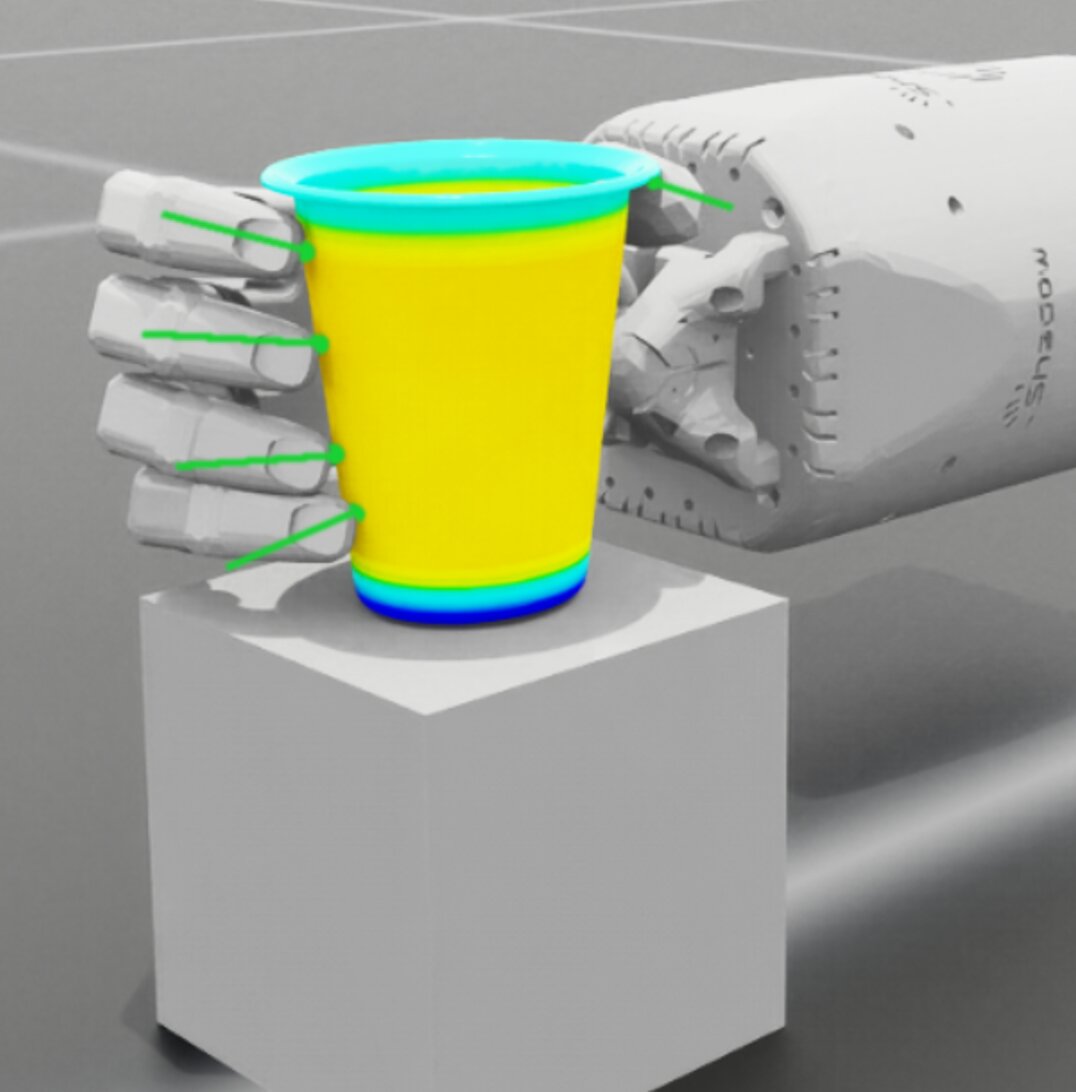}
  \hfill
  \includegraphics[width=0.32\columnwidth]{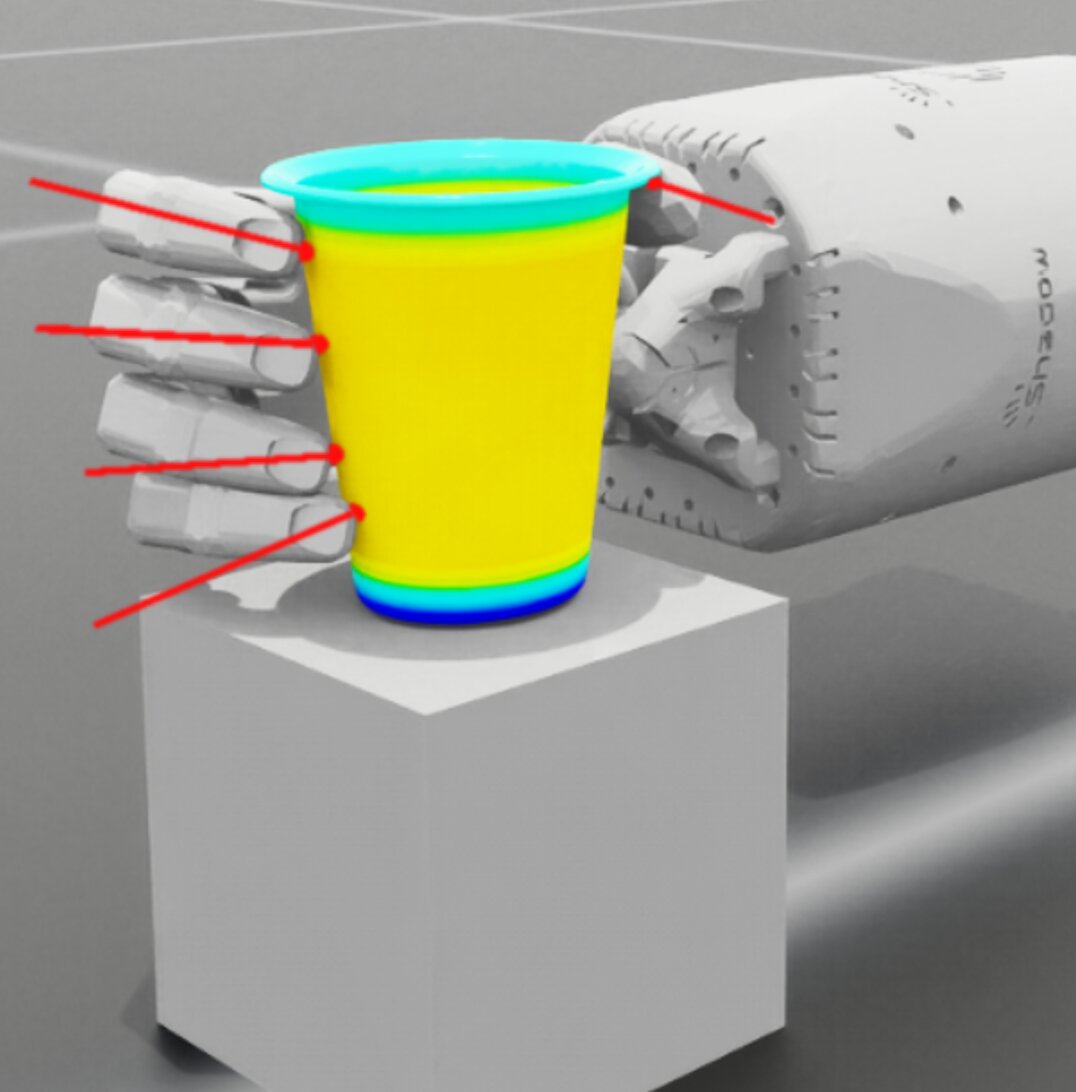}
  \hfill
  \includegraphics[width=0.32\columnwidth]{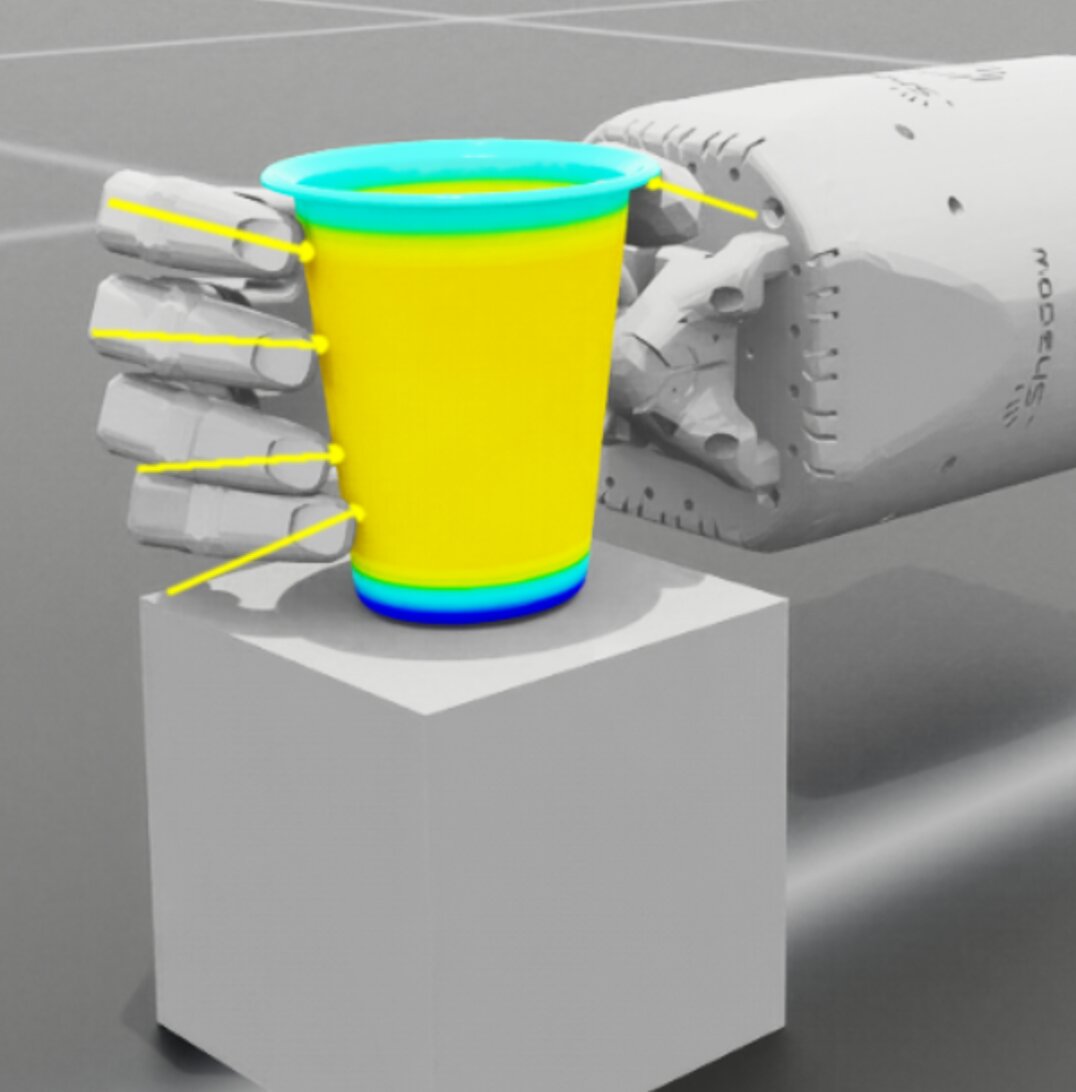}
  \caption{Contact force vectors for the three controller conditions.
  \textit{Left}: under-stiffness, object slips.
  \textit{Centre}: over-stiffness, red vectors indicate violations.
  \textit{Right}: proposed controller, all vectors green.}
  \label{fig:grip_forces}
\end{figure}

\begin{table}[h]
\caption{Grip Execution Results, Plastic Cup
($F^*_{\text{wall}}{=}0.3$~N).}
\label{tab:exp2}
\centering\small\setlength{\tabcolsep}{4.5pt}
\renewcommand{\arraystretch}{1.15}
\begin{tabular}{lcccc}
\hline
\textbf{Condition} & $\bm{f_{\max}}$ \textbf{[N]}
  & \textbf{Margin} & \textbf{Viol.} & \textbf{Success}\\
\hline
Under-stiffness        & 0.08 & -  & 0/5 & $\times$ \\
Over-stiffness         & 0.74 & 0.41 & 4/5 & $\times$ \\
Ours ($\lambda{=}0.7$) & 0.21 & 1.43 & 0/5 & \checkmark \\
\hline
\end{tabular}
\end{table}

The under-stiffness controller fails to retain the object; the
over-stiffness controller retains it but exceeds the admissible load
by $2.5\times$. The proposed controller achieves stable retention
(margin 1.43, no violations) and scales grip forces proportionally to
operator intent, with margin narrowing to 1.03 at \textit{"firmly"},
validating end-to-end propagation of the language command.
\begin{table}[h]
\caption{Effect of Interaction Mode $\lambda$ (proposed controller).}
\label{tab:lambda}
\centering\small\setlength{\tabcolsep}{5pt}
\renewcommand{\arraystretch}{1.15}
\begin{tabular}{lcccc}
\hline
\textbf{Mode} & $\bm{\lambda}$ & $\bm{f_{\max}}$ \textbf{[N]}
  & \textbf{Margin} & \textbf{Success}\\
\hline
Gently  & 0.3 & 0.09 & 3.33 & \checkmark \\
Default          & 0.7 & 0.21 & 1.43 & \checkmark \\
Firmly  & 1.0 & 0.29 & 1.03 & \checkmark \\
\hline
\end{tabular}
\end{table}

\section{Conclusions and Future Work}

We presented a pipeline that jointly solves grasp selection and grip
force regulation through a spatially distributed map of locally
admissible contact loads. The central contribution – force-map-aware
re-ranking – selects the grasp whose contacts fall in mechanically
admissible surface regions; an adaptive impedance controller then
scales per-finger stiffness to the locally admissible force.
Experiments on three objects demonstrate up to $12.3\times$
improvement in minimum admissible contact force over a geometry-only
baseline, zero force-bound violations, and controllable grip force
scaling with the operator's interaction mode – from margin 3.33 at
\textit{"gently"} to 1.03 at \textit{"firmly"}.

Future work will focus on deploying the pipeline on a physical robotic
hand with closed-loop tactile feedback, generalising to complex object
geometries, and developing goal-aware grasp generation that biases
sampling toward mechanically favorable regions from the outset.

\addtolength{\textheight}{-3cm}

\end{document}